\documentclass{sigplanconf}

\usepackage{amsmath}
\usepackage{graphicx}

\newcommand{\Eclipse} {ECL$^{i}$PS$^{e}$}
\newcommand{\sCOMMA} {\textsf{\footnotesize{s-COMMA}}}
\newcommand{\sCOMMAGUI} {\textsf{\footnotesize{s-COMMA GUI}}}

\newcommand{\flatsCOMMA} {\textsf{\footnotesize{Flat s-COMMA}}}
\newcommand{\code}[1]{\texttt{\small{#1}}}

\begin{document}

\conferenceinfo{PPDP'08,} {July 16--18, 2008, Valencia, Spain.}
\CopyrightYear{2008}
\copyrightdata{978-1-60558-117-0/08/07}


\title{Model-Driven Constraint Programming}

\authorinfo{Rapha\"el Chenouard}
           {CNRS, LINA, Universit\'e de Nantes, France.}
           {raphael.chenouard@univ-nantes.fr}

\authorinfo{Laurent Granvilliers}
           {CNRS, LINA, Universit\'e de Nantes, France.}
           {laurent.granvilliers@univ-nantes.fr}

\authorinfo{Ricardo Soto}
           {CNRS, LINA, Universit\'e de Nantes, France.\\
            Pontificia Universidad Cat\'olica de Valpara\'iso, Chile.}
           {ricardo.soto@univ-nantes.fr}


\maketitle

\begin{abstract}
Constraint programming can definitely be seen as a model-driven paradigm. The
users write programs for modeling problems. These programs are mapped
to executable models to calculate the solutions. This paper focuses on
efficient model management (definition and transformation). From this
point of view, we propose to revisit the design of
constraint-programming systems. A model-driven architecture is
introduced to map solving-independent constraint models to
solving-dependent decision models. Several important questions are
examined, such as the need for a visual high-level modeling
language, and the quality of metamodeling techniques to implement the
transformations. A main result is the \sCOMMA{} platform that
efficiently implements the chain from modeling to solving constraint
problems.
\end{abstract}




\section{Introduction}\label{sec:intro}

In constraint programming (CP), programmers define a model of a
problem using \emph{constraints} over \emph{variables}. The variables
may take values from domains, typically boolean, integer, or rational
values. The solutions to be found are tuples of values of the
variables satisfying the constraints. The search process is performed by
powerful solving techniques, for instance backtracking-like procedures and 
consistency algorithms to explore and reduce the space of 
potential solutions. In the past, CP has been shown to be efficient 
for solving hard combinatorial problems.

CP systems evolved from the early days of constraint logic programming
(CLP). In a CLP system, the constraint language is embedded in a logic
language, and the solving procedure combines the SLD-resolution with
calls to constraint solvers~\cite{JaffarPOPL1987}. The logic language
can be replaced with any computer programming language (e.g. C++ in 
ILOG Solver~\cite{PugetSCIS1994} or Java in Gecode/J~\cite{Gecode}) and even 
term rewriting~\cite{FruehwirthJLP1998}.
It turns out that the programming task may be hard, especially for non
experts of CP or computer programming. In this approach, modeling
concerns are not enough to write programs, and it is often mandatory
to deal with the encoding aspects of the host language or to
tune the solving strategy. In response to this problem, almost pure
modeling languages have been built, such as OPL~\cite{VanHentenryckBook1999} 
and Zinc~\cite{RafehPADL07}.

The design of the last generation of CP systems has been governed by
the idea of separating modeling and solving capabilities (e.g.
Essence~\cite{FrischIJCAI2007} and MiniZinc~\cite{Nethercote2007}). 
The system architecture has three layers, including the modeling language, 
the solvers, and a middle tool composed by a set of solver-translators 
implementing the mappings. In particular, this approach gives important benefits: The full 
expressiveness of CP is supported by a unique high-level modeling language, 
which is expected to be simple enough for non experts. The user is able 
to process one model with different solvers, a crucial feature for 
easy and fast problem experimentation. The platform is open to plug new solvers.

Our work follows this solver-independent idea, but under a Model-Driven 
Development (MDD) approach~\cite{OMG_MDA}, which is
well-known in the software engineering sphere. General requirements
have been defined for MDD architectures in order to define concise
models, to enable interoperability between tools, and to easily program 
mappings between models. The classical MDD infrastructure 
uses as base element the notion of a metamodel, which allows one 
to clearly define the concepts appearing in a model.

In this paper, the MDD approach is applied to a CP system. The goal is
to implement the chain from modeling to solving constraints. Our
approach is to transform user solving-independent models defined
through a visual modeling language to solver (executable) models using a
metamodeling strategy. CP concepts like domains, variables,
constraints, and relations between them are defined in a metamodel,
and thus the transformation rules are able to map these concepts from 
a source language to a target one. It results in a 
flexible and extensible architecture, robust enough to support changes at 
the mapping tool level. Moreover, we believe that the study of
metamodels for CP is of interest.

These ideas have been implemented in the \sCOMMA{}
platform~\cite{s-COMMA}. The front tool allows users to graphically
define constraint models. It is made on top of a general
object-oriented constraint language~\cite{SotoICTAI07}. Many solvers have
been plugged in the platform such as \Eclipse~\cite{Wallace97eclipse},
Gecode/J~\cite{Gecode}, GNU Prolog~\cite{DiazSAC00} and Realpaver~\cite{GranvilliersACM2006}. 
Upgrades are supported at the mapping tool, new solver-translators 
can be added by means of the AMMA platform~\cite{Kurtev2006}. 

The language for stating constraints in \sCOMMA{} is clearly not the novel part 
of the platform, in fact it includes typical and state-of-the-art modeling constructs 
and features. Novelty arises from the introduction of a solver-independent 
visual language -- which we believe is intuitive and simple enough for non experts --, 
and the use of a MDD approach involving metamodeling techniques to implement the mappings.

The outline of this paper is as follows. The MDD architecture
proposed is introduced in Section~\ref{sec:MDD-approach}. The \sCOMMA{} 
modeling language and the associated graphical interface are presented in
Section~\ref{sec:modeling-tool}. The mapping tool and the metamodeling
techniques used to develop solver-translators are explained in Section~\ref{sec:mapping-tool}. 
Some experimental results are then discussed in Section~\ref{sec:bench}. The related 
work and conclusion follows.

\section{A MDD approach for CP}\label{sec:MDD-approach}

Model-Driven Engineering (MDE) aims to consider models as first class
entities. A model is defined according to the semantics of a model of
models, also called a \textbf{metamodel}. A metamodel describes the concepts
appearing in a model, but also the links between these concepts, such as:
inheritance, composition or simple association.

\begin{figure}[!htbp]
\begin{center}
\includegraphics[width = 0.8\linewidth]{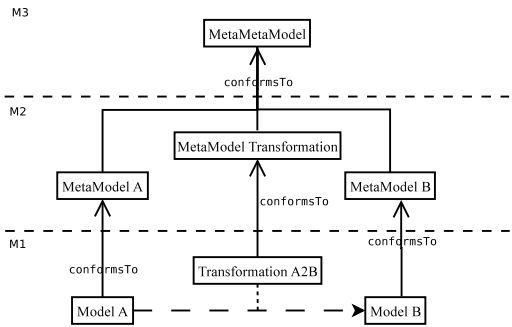}
\caption{A general MDA for model transformation. \label{fig:mda}}
\end{center}
\end{figure}

Figure \ref{fig:mda} depicts a general Model-Driven Architecture (MDA) for 
model transformation. Level M1 holds the model. Level 
M2 describes the semantic of the level M1 and thus identifies concepts 
handled by this model through a metamodel. Level M3 is the specification 
of level M2 and is self-defined. Transformation rules are defined to translate 
models from a source model to a target one, the semantic of these rules 
is also defined by a metamodel.

A major strength of using this metamodeling approach is that models are concisely
represented by metamodels. This allows one to define transformation rules that only 
operate on the concepts of metamodels (at the M2 level of the MDA approach), not on the
concrete syntax of a language. Syntax concerns are defined independently (we illustrate 
this in Section~\ref{sec:mapping-tool}). This separation is a great advantage for a 
clearly definition of transformation rules and grammar descriptions, which are the base of 
our mapping tool.

\begin{figure}[!htbp]
\begin{center}
\includegraphics[width = 0.95\linewidth]{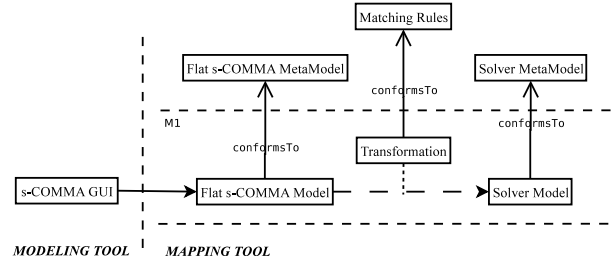}
\caption{The MDD architecture of s-COMMA. \label{fig:mdaSCOMMA}}
\end{center}
\end{figure}

Let us now illustrate how this approach is implemented in our platform.
Figure~\ref{fig:mdaSCOMMA} shows the MDD \sCOMMA{} architecture, which is 
composed by two main parts, a modeling tool and a mapping tool. 

The \sCOMMAGUI{} is our modeling tool, and it allow users 
to state constraint models using visual artifacts. An exactly textual representation of this 
language is also provided (for who does not
want to use visual artifacts). Both languages are solver-independent and are designed conform to the same 
metamodel (see Section~\ref{sec:modeling-tool}). The output of the \sCOMMAGUI{} is \flatsCOMMA{} an 
intermediate language which is still solver-independent but, in terms of abstraction is closer to the solver level. 
The goal is to simplify the development of solver-translators. \flatsCOMMA{} is also designed conform 
to a metamodel (see Section~\ref{sec:flatscomma}).

The mapping tool is composed by a set of solver-translators. Solver-translators are designed to 
match the metamodel concepts of \flatsCOMMA{} to the concepts of the solver metamodel
(see Section~\ref{sec:mapping-tool}). This process is defined conform to the general MDA for model 
transformation.

\vspace*{-4mm}
The \sCOMMAGUI{} is written in Java (about 30000 lines) and translators
are developed using the AMMA platform. The whole system allows to perform
the complete process from visual models to solver models. The system involves
several metamodels: an \sCOMMA{} metamodel, a \flatsCOMMA{} and solver metamodels.
The \sCOMMA{} metamodel has been built just for defining the concepts of the \sCOMMA{}
textual and visual language, it is not used to map \sCOMMA{} to \flatsCOMMA{}. 
For this task we already have an efficient translator. Our key aim of using metamodeling 
techniques is to provide an easier way to develop new solver-translators, compared
to the task of writing translators by hand.

In the following two sections we present the main parts of this architecture:
The modeling and the mapping tool, respectively.

\section{Modeling Tool}\label{sec:modeling-tool}

We have built our \sCOMMAGUI{} modeling tool on top of the \sCOMMA{} language. The \sCOMMA{}
language is defined through its metamodel and it has been designed to represent
the concepts of constraint problems, also called constraint satisfaction problems 
(CSPs). In this metamodel, the CSP concepts such as variables and domains have been 
merged with object-oriented concepts in order to state CSPs using an object-oriented 
style. The result is an object-oriented visual language for modeling CSPs. These 
decisions are supported by the following benefits:

\begin{itemize}

\item A problem is generally composed of several parts which may represent objects.
      They are naturally specified through classes.
      Thus, we obtain a more modular model, instead of forcing
      modelers to state the entire problem in a single block of code.

\item We gain similar benefits -- constraint and variable encapsulation, composition, inheritance, reuse -- 
      to those gained by writing software in a object-oriented programming language.

\item Visual artifacts are more intuitive to use and give a clearer view of the complete
      structure of the problem.

\end{itemize}

\begin{figure}[!htbp]
\begin{center}
\includegraphics[width = 0.7\linewidth]{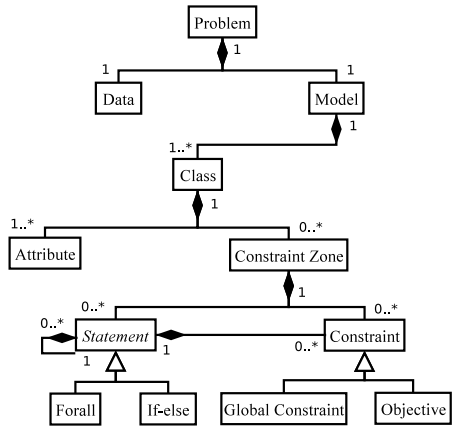}
\caption{s-COMMA Metamodel. \label{fig:scommametamodel}}
\end{center}
\end{figure}

Figure~\ref{fig:scommametamodel} illustrates the main concepts of the \sCOMMA{} metamodel
using UML class diagram notation. The role of each one of these concepts is explained in the 
following paragraphs. 

\subsection{s-COMMA models}

\begin{figure}[!htbp]
\begin{center}
\includegraphics[width=0.8\linewidth]{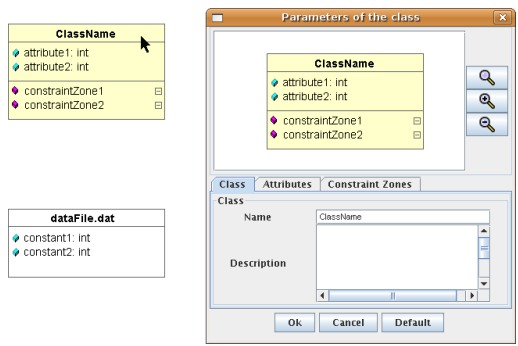}
\caption{Class artifact used in s-COMMA GUI\label{fig:classartifact}}
\end{center}
\end{figure}

The \sCOMMA{} metamodel defines the concepts appearing in \sCOMMA{} models. Thus, conform to this metamodel
an \sCOMMA{} model must be composed by two main parts, the model and data.
The model describes the structure of the problem and the data contain the constant 
values used by the model. In our \sCOMMAGUI{} front tool this problem's 
structure is represented by class artifacts and the data concept is represented by the data 
artifact\footnote{Artifacts used on the \sCOMMAGUI{} have been adapted from the class artifact provided by the UML 
Infrastructure Library Basic Package. This 
adaptation is completely allowed by the UML Infrastructure 
Specification~\cite{OMG_UML}.} (see Figure~\ref{fig:classartifact}). 

\subsubsection{Class artifacts}

Class artifacts have by default three compartments, the upper compartment for the class 
name, the middle compartment for attributes and the bottom one for constraint 
zones. By clicking on the class artifact its specification can be opened to define 
its class name, their attributes and constraint zones. Relationships can be used to 
define inheritance (a subclass inherits all attributes and constraint zones of its superclass) or 
composition between classes.

\subsubsection{Data artifacts}

Data artifacts have two compartments, one for the file name and another for both the 
constants and variable-assignments. Constants, also called data variables can be defined with a real, integer or 
enumeration type. Arrays of one dimension and arrays of two dimensions of constants are allowed. 
Variable-assignment corresponds to the assignment of a value to a variable of an object. Variable-assignments 
can also be performed if objects are inside an array (see an example in Section~\ref{sec:stablemarriage}).

\subsubsection{Attributes}

Attributes may represent decision variables, sets, objects or arrays. Decision 
variables can be defined by an integer, real or boolean type. Sets can be composed of
integers or enumeration values. Objects are instances of 
classes which must be typed with their class name. Arrays of one and two dimensions are allowed, 
they can contain decision variables, sets or objects. Decision variables, sets and arrays can be 
constrained to a determined domain.

\subsubsection{Constraint Zones}

Constraint zones are used to group constraints encapsulating them inside a class. 
A constraint zone is stated with a name and it can contain the following elements:

\begin{itemize}

\item \textit{Constraints:} Typical operations and relations are provided 
to post constraints. For example, comparison relations (\code{<,>,} 
\code{<=,>=,=,<>}), arithmetic operations (\code{+,*,} \code{-,/}), logical relations 
(\code{and,or,xor,} \code{->,<-,<->}), and set operations (\code{in, subset, 
superset, union, diff, symdiff, intersection, cardinality}).

\item \textit{Statements:} Forall and conditional statements are supported. The forall
(e.g. \code{forall(i in 1..5)}) is stated by declaring a loop-variable (\code{i}) 
and the set of values to be traversed (\code{1..5}). A loop-variable is a local 
variable and it is valid just inside the loop
where it was declared. Conditionals are stated by means of if-else expressions. 
For instance, \code{if(a) b else c;} where \code{a} is the condition, which can 
includes decision variables; and \code{b} and \code{c} are the alternatives, 
which may be statements or constraints. 

\item \textit{Objective:} objective functions are allowed and they can be stated 
by tagging the expression involved with the selected option (e.g. \code{[minimize] x+y+z;}).

\item \textit{Global Constraints}: a basic set of global constraints 
(e.g. alldifferent, cumulatives) is supported. Additional constraints
can be integrated to this basic set by means of extension mechanisms 
(for details refer to~\cite{SotoICTAI07}).

\end{itemize}

\subsection{The stable marriage problem}\label{sec:stablemarriage}

Let us now illustrate some of these concepts in the \sCOMMAGUI{} by means of
the stable marriage problem. 

\begin{figure}[!htbp]
\begin{center}
\includegraphics[width=1\linewidth]{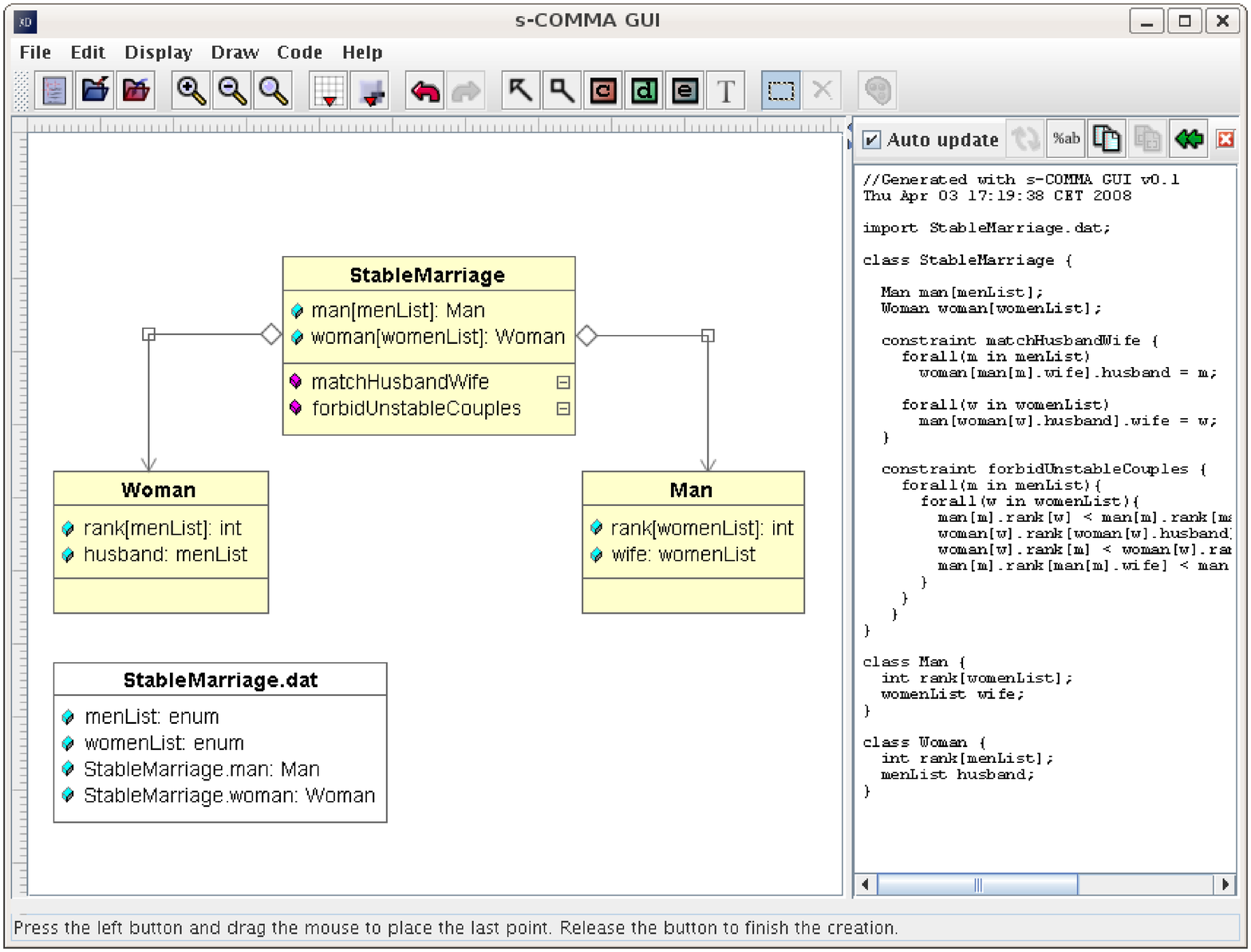}
\caption{The stable marriage problem on the s-COMMA GUI\label{fig:GUI}}
\end{center}
\end{figure}

Consider a group of $n$ women and a 
group of $n$ men who must marry. Each women has 
a preference ranking for her possible husband, and each men 
has a preference ranking for his possible wife. The problem is to find a 
matching between the groups such that the marriages are stable i.e., there are 
no pair of people of opposite sex that like each other better than their
respective spouses. 

Figure~\ref{fig:GUI} shows a snapshot of the \sCOMMAGUI{} where the stable
marriage problem is represented by a class diagram. This diagram is composed by three 
classes, one class to represent men, one to represent women, and a main class 
to describe the stable marriages. Once the user states a visual artifact, 
the corresponding \sCOMMA{} textual version is automatically generated on the 
right-panel of the tool. For readability we illustrate the textual version of
the problem in Fig.~\ref{fig:stablemarriage}

\begin{figure}[!htbp]
\begin{scriptsize}
\begin{verbatim}
//Model file                                        
1. import StableMarriage.dat;                          
2.                                                   
3. class StableMarriage {                          
4.                                                  
5.   Man man[menList];                              
6.   Woman woman[womenList];                        
7.                                                   
8.   constraint matchHusbandWife {                  
9.     forall(m in menList)                         
10.      woman[man[m].wife].husband = m;
11.     
12.    forall(w in womenList) 
13.      man[woman[w].husband].wife = w;
14.  }
15.
16.  constraint forbidUnstableCouples {
17.    forall(m in menList){
18.      forall(w in womenList){
19.        man[m].rank[w] < man[m].rank[man[m].wife] -> 
20.        woman[w].rank[woman[w].husband] < woman[w].rank[m];
21.
22.        woman[w].rank[m] < woman[w].rank[woman[w].husband] -> 
23.        man[m].rank[man[m].wife] < man[m].rank[w];
24.      } 
25.    }  
26.  }  
27.}
28. 
29. class Man { 
30.   int rank[womenList]; 
31.   womenList wife; 
32. }
33.
34. class Woman {
35.   int rank[menList];
36.   menList husband;
37. } 

//Data file
1. enum menList := {Richard,James,John,Hugh,Greg};
2. enum womenList := {Helen,Tracy,Linda,Sally,Wanda};
3. Man StableMarriage.man := 
    [Richard: {[Helen:5 ,Tracy:1, Linda:2, Sally:4, Wanda:3],_},
     James  : {[Helen:4 ,Tracy:1, Linda:3, Sally:2, Wanda:5],_},
     John   : {[Helen:5 ,Tracy:3, Linda:2, Sally:4, Wanda:1],_},
     Hugh   : {[Helen:1 ,Tracy:5, Linda:4, Sally:3, Wanda:2],_},
     Greg   : {[Helen:4 ,Tracy:3, Linda:2, Sally:1, Wanda:5],_}];

4. Woman StableMarriage.woman := 
    [Helen: {[Richard:1, James:2, John:4, Hugh:3, Greg:5],_},
     Tracy: {[Richard:3, James:5, John:1, Hugh:2, Greg:4],_},
     Linda: {[Richard:5, James:4, John:2, Hugh:1, Greg:3],_},
     Sally: {[Richard:1, James:3, John:5, Hugh:4, Greg:2],_},
     Wanda: {[Richard:4, James:2, John:3, Hugh:5, Greg:1],_}];
\end{verbatim}
\end{scriptsize}
\caption{An s-COMMA model for the stable marriage problem.}
\label{fig:stablemarriage}
\end{figure}

The class representing men (at line 29 in the model file) is composed by one array containing 
integer values which represents the preferences of a man, the array is indexed by the enumeration type \code{womenList} 
(at line 2 in the data file), thereby the 1st index of the array is \code{Helen}, the 2nd is 
\code{Tracy}, the third is \code{Linda} and so on. Then, an attribute called \code{wife} is 
defined (line 31), which represents the spouse of an object man. This variable has \code{womenList} as a 
type which means that its domain is given by the enumeration \code{womenList}. The definition 
of the class \code{Women} is analogous. 

The class \code{StableMarriage} has a more complex declaration. We first define two arrays,
one called \code{man} which contains objects of the class \code{Man} and other which 
contains objects of the class \code{Woman}. Each one represents the group of men and the group of women, 
respectively. The composition relationship between classes can be seen on the class diagram.

At line 8 a constraint zone called \code{matchHusbandWife} is stated. In this
constraint zone, two \code{forall} loops including a constraint are posted  
to ensure that the pairs man-wife match with the pairs woman-husband. 
The \code{forbidUnstableCouples} constraint zone contains two loops 
including two logical formulas to ensure that marriages are stable.

The data file is called by means of an import statement (at line 1). This file contains two enumeration 
types, \code{menList} and \code{womenList}, which have been used in the model as a type, for indexing
arrays, and as the set of values that loop-variables must traverse. \code{StableMarriage.man} is a
variable-assignment for the array called \code{man} defined at line 5 in the model file. 

This variable-assignment
is composed by five objects (enclosed by \code{`$\lbrace\rbrace$'}), one for each men of the group. 
Each of these objects has two elements, the first element\footnote{Let us note that we use standard modeling 
variable-assignments, that is, assignments are performed respecting the order of the class' 
attributes: the first element of the variable-assignment is matched with the 
first attribute of the class, the second element of the variable-assignment with the 
second attribute of the class and so on.} is an array (enclosed by \code{`[ ]'}). This array 
sets the preferences of a men, assigning the values to the array \code{rank} of a \code{Man} object (e.g. Richard
prefers Tracy 1st, Linda 2nd, Wanda 3rd, etc).

The second element is an underscore symbol ('\_'). This symbol is used to omit assignments, so the 
variable \code{wife} remains as a decision variable of the problem i.e., a variable for which the solver 
must search a solution.

\subsection{Flat s-COMMA models}\label{sec:flatscomma}

Before explaining how \sCOMMA{} models are mapped to their equivalent solver
models, let us introduce the intermediate \flatsCOMMA{} language.

\begin{figure}[!htbp]
\begin{center}
\includegraphics[width = 0.6\linewidth]{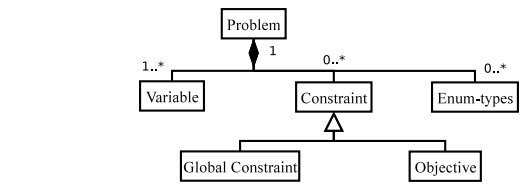}
\caption{Flat s-COMMA Metamodel. \label{fig:flatscommametamodel}}
\end{center}
\end{figure}

\flatsCOMMA{} has been designed to simplify the transformation process from
\sCOMMA{} models to solver models. In \flatsCOMMA{} much of the constructs
supported by \sCOMMA{} are transformed to simpler ones, in order to
be closer to the form required by classical solver languages. \flatsCOMMA{}
is also defined by a metamodel.

Figure~\ref{fig:flatscommametamodel} illustrates the main elements of the \flatsCOMMA{}
metamodel, where many \sCOMMA{} concepts have been removed. Now, the 
metamodel is mainly a definition of a problem composed by variables (decision 
variables) and constraints. 

In order to transform \sCOMMA{} to \flatsCOMMA{}, several steps are 
involved, which are explained in the following.

\begin{itemize}

\item Enumeration substitution: In general solvers do not support non-numeric types. So,
enumerations are replaced by integer values. However, enumeration values are stored 
to show the results in the correct format.

\item Data substitution: Data variables stated in the model file are replaced by their 
corresponding values i.e., the value defined in the data file. 

\item Loop unrolling: Loops are not widely supported by solvers, hence we generate 
an unrolled version of the forall loop. 

\item Flattening composition: The hierarchy generated by composition is 
flattened. This process is done by expanding each object declared in the main 
class adding its attributes and constraints in the \flatsCOMMA{} file. The name 
of each attribute has a prefix corresponding to the concatenation 
of the names of objects of origin in order to avoid name redundancy. 

\item Conditional removal: Conditional statements are transformed to logical formulas.
For instance, \code{if a then b else c} is replaced by $(a\Rightarrow b) \wedge (a \vee c)$.

\item Logic formulas transformation: Some logic operators are not supported by 
solvers. For example, logical equivalence ($a \Leftrightarrow b$) and reverse 
implication ($a \Leftarrow b$). We transform logical equivalence expressing it in 
terms of logical implication ($(a \Rightarrow b) \wedge (b \Rightarrow a)$). Reverse 
implication is simply inverted ($b \Rightarrow a$).
\end{itemize}

\begin{figure}[htbp]
\begin{scriptsize}
\begin{verbatim}
1.  variables: 
2.
3.    womenList man_wife[5] in [1,5];
4.    menList woman_husband[5] in [1,5];
5.
6.  constraints: 
7.
8.    woman_husband[man_wife[1]]=1;
9.    woman_husband[man_wife[2]]=2;
10.   woman_husband[man_wife[3]]=3;
11.   ...
12.
13.   man_wife[woman_husband[1]]=1;
14.   man_wife[woman_husband[2]]=2;
15.   man_wife[woman_husband[3]]=3;
16.   ...
17.    
18.   5<man_1_rank[man_wife[1]] -> 
19.   woman_1_rank[woman_husband[1]]<1;
20.   1<woman_1_rank[woman_husband[1]] -> 
21.   man_1_rank[man_wife[1]]<5;
22.
23.   1<man_1_rank[man_wife[1]] -> 
24.   woman_2_rank[woman_husband[2]]<3;
25.   3<woman_2_rank[woman_husband[2]] -> 
26.   man_1_rank[man_wife[1]]<1;
27.   ...
28.
29. enum-types:
30.
31.   menList := {Richard,James,John,Hugh,Greg};
32.   womenList := {Helen,Tracy,Linda,Sally,Wanda};
\end{verbatim}
\end{scriptsize}
\caption{The Flat s-COMMA model of the stable marriage problem.}
\label{fig:flatstablemarriage}
\end{figure}

Figure~\ref{fig:flatstablemarriage} depicts the \flatsCOMMA{} model of the stable marriage problem. The file is composed
of two main parts, variables and constraints. Variables at lines 3-4 are generated by the
flattening composition process. The array \code{man} composed by objects of type \code{Man}
is decomposed and transformed to a single array of decision variables. The array 
\code{man\_wife} contains the decision variables \code{wife} of the original 
array \code{man}; and the array \code{woman\_husband} contains the decision variables 
\code{husband} of the original array \code{woman}. The arrays rank of both objects
\code{Man} and \code{Woman} are not considered as decision variables since they have 
been filled with constants (at lines 3-4 of the data file in Figure~\ref{fig:stablemarriage}).
The size of the array \code{man\_wife} is 5, this value is given by the enumeration
substitution step which sets the size of the array with the size of the enumeration 
\code{menList} (\code{5}). The domain \code{[1,5]} is also given by this step which 
states as domain an integer range corresponding to the number of elements of the enumeration 
used as a type (\code{womenList}) by the attribute \code{wife}. 
The type of both arrays is maintained to give the solutions 
in the enumeration format. These values are stored in the block \code{enum-types}. Lines 8-15 
come from the loop unrolling phase of the forall statements of the
\code{matchHusbandWife} constraint zone. Likewise, lines 18-26 are generated by the loops
of \code{forbidUnstableCouples}. In these constraints, the data substitution step has replaced
several constants with their corresponding integer values.

%

\section{Mapping Tool}\label{sec:mapping-tool}

In this section we explain the mechanisms provided by the MDD approach to develop 
our solver-translators. These translators are designed to perform the mapping 
from \flatsCOMMA{} to solver models. We use the AMMA platform as our base tool to 
build them.

The AMMA platform allows one to develop this task by means of two 
languages: KM3~\cite{Jouault2006KM3} and ATL~\cite{Jouault2006ATL}. KM3 is used to define metamodels, and ATL is used
to describe the transformation rules and also to generate the target file. 

\subsection{KM3}\label{sec:KM3}

The Kernel Meta Meta Model (KM3) is a language to define metamodels. 
KM3 has been designed to support most metamodeling standards and
it is based on the simple notion of classes to define each one of the 
concepts of a metamodel. These concepts will then be used by the 
transformation rules and to generate the target file. 
Figure~\ref{fig:flatscommaKM3} illustrates an extract of the \flatsCOMMA{} 
metamodel written in KM3. 

\begin{figure}[!htbp]
\begin{scriptsize}
\begin{verbatim}
1.  class Problem {
2.    attribute name : String;
3.    reference variables[1-*] container : Variable;
4.    reference constraints[0-*] container : Constraint;
5.    reference enumTypes[0-*] container : EnumType; 
6.  }	
7.	
8.  class Variable {
9.    attribute name : String;
10.   attribute type : String;
11.   reference array [0-1] container : Array;
12.   reference domain container : Domain;
13. }
14.  
15. class Array {
16.   attribute row : Integer;
17.   attribute col[0-1] : Integer;
18. }
\end{verbatim}
\end{scriptsize}
\caption{An extract of the Flat s-COMMA KM3 metamodel. \label{fig:flatscommaKM3}}
\end{figure}

The \flatsCOMMA{} KM3 metamodel states that the concept \code{Problem}
is composed of one attribute and three references. The attribute \code{name}
at line 2 represents the name of the model and it is declared with the basic 
type \code{String}. Line 3 simply states that the class \code{Problem} is
composed by a set of objects of the class \code{Variable}. The reserved 
word \code{reference} is used to declare links with instances of other classes 
and the statement \code{[1-*]} defines the multiplicity of the relationship.
If the multiplicity statement is omitted the relationship is defined as 
\code{[1-1]}. Lines 4-5 are similar and define that the class \code{Problem}
is also composed by \code{constraints} and \code{enumTypes} (values
stored by the enumeration substitution step). Remaining classes are defined
in the same way.

\subsection{ATL}\label{sec:ATL}

The Atlas Transformation Language (ATL) allow us to
define transformation rules according to one or several metamodels.
The rules clearly state how concepts from source metamodels
are matched to concepts of the target ones. Figure~\ref{fig:rulesATL}
shows some of the ATL rules used to transform the concepts
of the \flatsCOMMA{} metamodel to the concepts of the
Gecode/J metamodel. The metamodel of Gecode/J is not presented
here since it is very close to the \flatsCOMMA{} metamodel.
Indeed, most CP solver languages are used to
express quite the same concepts and \flatsCOMMA{} is designed to
be as close as possible from the solving level. This is a great
asset because transformation rules become simple: we mainly need
one to one transformations.

\begin{figure}[htbp]
\begin{scriptsize}
\begin{verbatim}
1. module FlatsComma2GecodeJ;
2. create OUT : GecodeJ from IN : FlatsComma;
3. 
4. rule Problem2Problem {
5.   from
6.     s : FlatsComma!Problem (
7.     )
8.   to
9.     t : GecodeJ!Problem(
10.      name <- s.name,
11.      variables <- s.variables,
12.      constraints <- s.constraints,
13.      enumTypes <- s.enumTypes
14.    )
15. }
16.
17. rule Variable2Variable {
18.   from
19.     s : FlatsComma!Variable (
20.    not s.isArrayVariable
21.   )
22.  to
23.   t : GecodeJ!Variable (
24.    name <- s.name,
25.    type <- s.type,
26.    domain <- s.domain
27    )
28. }
29.
30. helper context FlatsComma!Variable def: 
31. isArrayVariable : Boolean=
32.   not self.array.oclIsUndefined();
\end{verbatim}
\end{scriptsize}
\caption{ATL rules for transformation from Flat s-COMMA to Gecode/J.\label{fig:rulesATL}}
\end{figure}

The first line of this file specifies the name of the
transformation. A \code{module} is used to define and
regroup a set of rules and helpers. Rules define the mappings,
and helpers allow to define factorized ATL code that can 
be called from different points of the ATL file (they can be viewed as 
the ATL equivalent to Java methods).

Line 2 states the target (\code{create}) and source metamodels (\code{from}).
The first rule presented is called \code{Problem2Problem} and defines 
the matching between the concepts \code{Problem} expressed in \flatsCOMMA{} and Gecode/J. 
The source elements are stated with the reserved word 
\code{from} and the target ones with the reserved word \code{to}.
These elements are declared like variables with a name (\code{s,t}) and 
a type corresponding to a class in a metamodel (\code{FlatsComma!Problem, 
GecodeJ!Problem}). In the target part of the rule the name attribute of 
the \flatsCOMMA{} problem is assigned to the Gecode/J name 
(\code{name <- s.name}), this is just an string assignment. However,
the following two statement are assignments between concepts that are
defined as \code{reference} in the metamodel. So, they need
a specific rule to carry out the transformation. For instance, the \flatsCOMMA{} KM3 
metamodel defines that the reference \code{variables} is composed by a set of 
\code{Variable} elements. Thus, the statement (\code{variables <- s.variables})
calls implicitly the rule \code{Variable2Variable}, which defines the match
between each element of objects \code{Variable}. It can be highlighted that the ATL
engine requires a unique name for each rule and a unique matching case: \code{from} and \code{to}
blocks. When several rules can be applied a guard (the boolean test in line 20) over 
the from statement must remove choice ambiguities.

The \code{Variable2Variable} rule matches three elements. The first two statements are
simple string assignments and the last one is a reference assignment. Let us remark that
a second rule to process array variables has been defined (but not presented here) which 
includes an additional statement for the array element. These two rules are distinguished
according to complementary guards over the source block using the helper
\code{isArrayVariable}. Guards act as filter on the source variable instances to
process. The presented helper \code{isArrayVariable} applies on variable instances
in \flatsCOMMA{} models and returns true when the instance contains an array
element. ATL inherits from OCL~\cite{OMG_OCL} syntax and semantics; and most OCL functions
and types are available within ATL. 

Although the rules used here are not complex, ATL is able to perform more difficult rules.
For instance, the most difficult rule we defined, was the transformation rule from \flatsCOMMA{} 
matrix containing sets, which must be unrolled in the \Eclipse{} models (since set matrix 
are not supported). This unroll process is carried out by defining a single set in \Eclipse{} 
for each cell in the matrix. The name of each single variable is composed by the name of the matrix, 
and the corresponding row and column index.

\begin{figure}[!htbp]
\begin{scriptsize}
\begin{verbatim}
1. rule Problem2Problem {
2.    from
3.      s : FlatsComma!Problem (
4.          s.hasSetMatrix
5.      )
6.    to
7.      t : ECLiPSe!Problem (
8.          name <- s.name,
9.          constraints <- s.constraints,
10.         enumTypes <- s.enumTypes
11.     )
12.   do {
13.     t.variables <- s.variables->collect(e|
14.      if e.isSetMatrix() then
15.        thisModule.getMatrixCells(e)->collect(f|
16.         thisModule.SetMatrixVariable2Variable(f.var,f.i,f.j)
17.        )
18.      else
19.        e
20.      endif
21.      )->flatten();
22.   }
23. }
24.
25. rule SetMatrixVariable2Variable(var : FlatsComma!Variable,
26.                                  i : Integer, j : Integer) {
27.   to
28.     t : ECLiPSe!Variable(
29.       name <- var.name + i.toString() + '_' + j.toString(),
30.       type <- var.type,
31.       domain <- var.domain,
32.     )
33.   do {
34.     t;
35.   }
36. }
\end{verbatim}
\end{scriptsize}
\caption{ATL rules for decomposing matrix containing sets.\label{fig:setmat}}
\end{figure}

Figure~\ref{fig:setmat} shows the rule \code{Problem2Problem} defined for \Eclipse{}, this rule has a condition
(line 4) to check whether set matrix are defined in the model. If the condition is true, \code{name}, \code{constraints} and \code{enumTypes} are 
matched normally, but \code{variables} has a special procedure to decompose the set matrix. 

This procedure begins at line 12 with a \code{do} block.
In this block, the \code{collect} loop iterates over the variables. Then, each of these variables (\code{e}) is checked
to determine whether it has been defined as a set matrix (line 14). 
If this occurs, the helper \code{getMatrixCells(e)} calculates the set 
of tuples corresponding to all the cells of the matrix (\code{thisModule} is used to call explicitly
helpers or rules). Each tuple is composed of the
\flatsCOMMA{} variable (\code{f.var}), a row index 
(\code{f.i}) and a column index (\code{f.j}). Then, the rule
\code{SetMatrixVariable2Variable} is applied to each tuple in order to
generate the \Eclipse{} variables. This rule does not contain a source block since the
source elements are the input parameters. The rule sets to the attribute \code{name}, the concatenation of the name of
the matrix with the respective row (\code{i.toString()}) and column (\code{j.toString()}).
Attributes \code{type} and \code{domain} are also matched. Finally, \code{flatten()} is an OCL inherited method
used to match the generated set of variables with \code{t.variables}.

ATL is also used to generate the solver target file. This is possible by defining a new ATL file
(called generically ATL2Text) where we can embed the concepts of the metamodel in the syntax
of the target file. This is done by means of a querying facility that enables to specify requests
onto models. 

\begin{figure}[!htbp]
\begin{scriptsize}
\begin{verbatim}
1.  query GecodeJ2Text = GecodeJ!Problem.allInstances()->
2.    asSequence()->first().toString2().
3.    writeTo('./GecodeJ/Samples/'+ thisModule.getFileName() +
4.    '.java');
5.
6.  helper context GecodeJ!Problem def: toString2() : String=
7.    'package comma.solverFiles.gecodej;\n' +  
8.    'import static org.gecode.Gecode.*;\n' +
9.    'import static org.gecode.GecodeEnumConstants.*;\n' +
10.   ...
11.
12.    self.variables->collect(e | e.toString2())
13.   ->iterate(e; acc:String = '' | acc +'    '+e) +
14.  ...
15.    '}\n\n'
16.  ;
17.
18. helper context GecodeJ!Variable def: toString2() : 
19.   String=
20.   if self.array.oclIsUndefined() then
21.     'IntVar ' + self.name + ' = new IntVar(this,\"' + 
22.     self.name + '\",' + self.domain.toString2() +');\n' +
23.     '    vars.add('+ self.name +');\n'
24.   else if self.array.col.oclIsUndefined() then
25.     'VarArray<IntVar> ' + self.name + ' = initialize(\"' + 
26.     self.name + '\",' + self.array.toString2() +
27.     ',' + self.domain.toString2()+');\n' + 
28.     '    vars.addAll(' + self.name + ');\n'
29.   else
30.     'VarMatrix<IntVar> ' + self.name + ' = initialize(\"' + 
31.     self.name + '\",' +  self.array.toString2() +
32.     ',' + self.domain.toString2()+');\n' + 
33.     '    vars.addAll(' + self.name + ');\n'
34.   endif endif
35. ;
\end{verbatim}
\end{scriptsize}
\caption{GecodeJ2Text file\label{fig:ATLtoText}}
\end{figure}

Figure~\ref{fig:ATLtoText} shows a fragment of the GecodeJ2Text definition 
to generate the Gecode/J file. Lines 1-4 states the query
on the \code{Problem} concept and defines the target file. Queries are able to call
helpers, which allow us to build the string to be written in the target solver file. This query 
calls the helper \code{toString2()} defined for the concept \code{Problem}. This helper
is stated at line 6 and it creates first the string corresponding to the headers (package and
import statements) of a Gecode/J model. Then, at lines 12-13 the string corresponding
to the variables declarations is created. This is done by iterating the collection of 
variables and calling the corresponding \code{toString2()} helper for the \code{Variable}
instances. This helper is declared at line 18, it defines three possible variable declarations, 
single variable (\code{IntVar}), a one dimension array (\code{VarArray<IntVar>}), and a two dimension
array (\code{VarMatrix<IntVar>}). The alternatives are chosen by means of an if-else statement.
The condition \code{self.array.oclIsUndefined()} checks whether the concept array is undefined. If this 
occurs, the variable corresponds to a single variable. The string representing this declaration
uses \code{self.name} which refers to the name of the variable, \code{self.domain.toString2()}
calls a helper to get the string representing the domain of the variable. The next alternative
tests if the attribute \code{col} of the \code{array} is undefined, in this case the variable is a 
one dimension array, otherwise it is a two dimension array. The call \code{self.array.toString2()} is used
in the two last alternatives, it returns the string corresponding to the size of arrays.

Figure~\ref{fig:stablegecodej} depicts an extract of the Gecode/J file generated for the stable 
marriage problem. Lines 1-3 states the headers. Line 6 declares the array called \code{man\_wife}.
which is initialized with size \code{5} and domain \code{[1,5]}. At line 8 the array is added
to a global array called \code{vars} for performing the labeling process. Lines 14-19 illustrate 
some constraints, which are stated by means of the \code{post} method.

\begin{figure}[htbp]
\begin{scriptsize}
\begin{verbatim}
1.  package comma.solverFiles.gecodej;
2.  import static org.gecode.Gecode.*;
3.  import static org.gecode.GecodeEnumConstants.*;
4.  ...
5.
6.  VarArray<IntVar> man_wife  = 
7.      initialize("man_wife",5,1,5);
8.  vars.addAll(man_wife);
9.
10. VarArray<IntVar> woman_husband  = 
11.      initialize("woman_husband",5,1,5);
12. vars.addAll(woman_husband);
13.
14. post(this, new Expr().p(get(this,woman_husband,
15.   get(man_wife,1))),IRT_EQ, new Expr().p(1));
16. post(this, new Expr().p(get(this,woman_husband,
17.   get(man_wife,2))),IRT_EQ, new Expr().p(2));
18. post(this, new Expr().p(get(this,woman_husband,
19.   get(man_wife,3))),IRT_EQ, new Expr().p(3));
20. ...
\end{verbatim}
\end{scriptsize}
\caption{Gecode/J model for the stable marriage problem.\label{fig:stablegecodej}}
\end{figure}

\subsection{TCS}\label{sec:TCS}

TCS~\cite{Jouault2006TCS} (Textual Concrete Syntax) is another language provided by the AMMA platform. 
TCS is not mandatory to add a new translator but it is involved in the process since it 
is the language used to parse the \flatsCOMMA{} file. TCS is able to perform
this task by bridging the \flatsCOMMA{} metamodel with the \flatsCOMMA{} grammar.

\begin{figure}[!htbp]
\begin{scriptsize}
\begin{verbatim}
1.  template Problem
2.   : "variables" ":" variables
3.     "constraints" ":" constraints
4.     "enum-types" ":" enumTypes
5.   ;
6.
7.  template Variable
8.   :  type name (isDefined(array) ? array) "in" domain ";"
9.   ;
10.
11. template Array
12.  : "[" row (isDefined(col) ? "," col ) "]"
13.  ;
\end{verbatim}
\end{scriptsize}
\caption{TCS for Flat s-COMMA.\label{fig:TCS}}
\end{figure}
 
Figure \ref{fig:TCS} shows an extract of the TCS file for \flatsCOMMA{}. Each 
class of the \flatsCOMMA{} metamodel has a dedicated template declared with 
the same name. Within templates, words between double quotes are tokens in the 
grammar (e.g. \code{"variables"}, \code{":"}). Words without double quotes are used 
to introduce the corresponding list of concepts. For instance \code{variables} is
defined as a reference to objects \code{Variable} in the class \code{Problem} of the metamodel. Thus,
\code{variables} is used to call their associate template i.e., the
\code{Variable} template. This template defines the syntactic structure
of a variable declaration. It has a conditional structure (\code{(isDefined(array) ? array)}),
which means that the template \code{Array} is only called if the variable is defined
as an array.

\subsection{Transformation process}\label{sec:trans-process}

\begin{figure}[!hbtp]
\begin{center}
\includegraphics[width=1\linewidth]{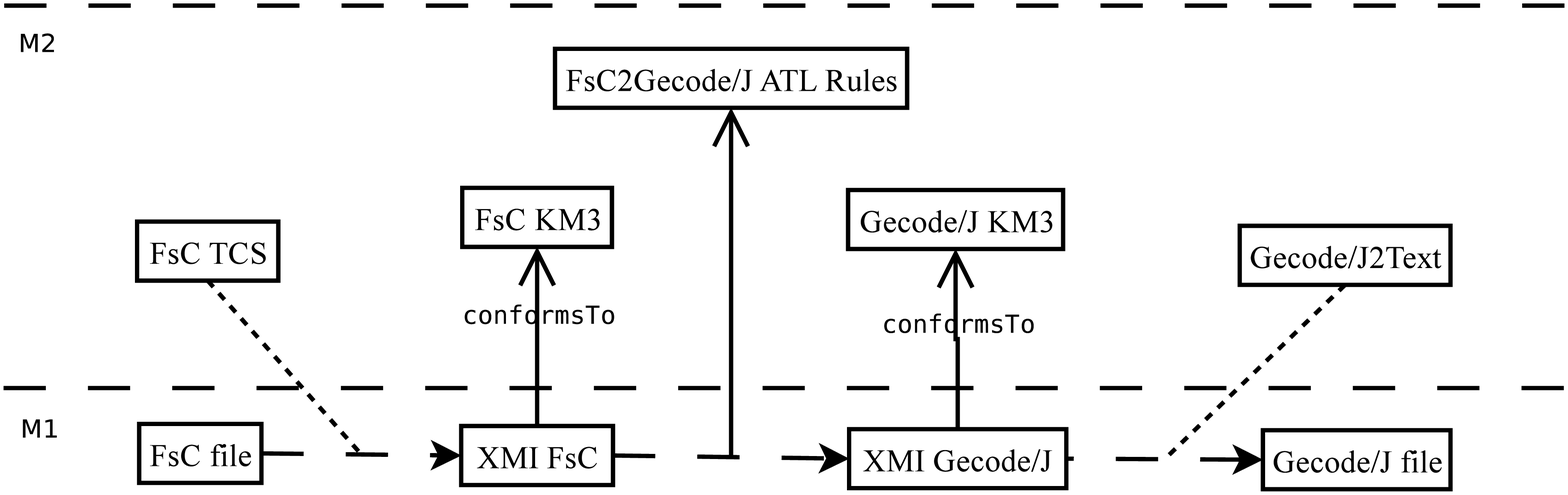}
\caption{The AMMA model-driven process on the example of Flat s-COMMA (FsC) to Gecode/J.\label{fig:AMMA}}
\end{center}
\end{figure}

TCS and KM3 work together and their compilation generates a Java 
package (which includes lexers, parsers and code generators) for \flatsCOMMA{} (FsC), 
which is then used by the ATL files to generate the target model.
Figure~\ref{fig:AMMA} depicts the complete transformation process.
The \flatsCOMMA{} file is the output of the \sCOMMAGUI{}, this file
is taken by the Java package which generates a XMI \footnote{XMI is 
the standard used for exchanging metadata in MDD architectures.} (XML Metadata Interchange) for 
\flatsCOMMA{}, this file includes an organized representation of models
in terms of their concepts in order to facilitate the task of transformation
rules. Over this file ATL rules act and generate a XMI file for Gecode/J. 
Finally this file is taken by the Gecode/J2Text which builds the solver file.

The complete process involves TCS, KM3 and ATL. But, the integration of a new 
translator just requires KM3 and ATL (the mapping tool only needs one TCS 
file). As we mention in Section~\ref{sec:ATL}, solver metamodels are 
almost equivalents, and ATL rules are mainly one to one mappings. 
As a consequence, the development of KM3 and ATL rules for new solver-translators 
should not be a hard task. So, we could say that the concrete work for plugging a new 
solver is reduced to the definition of the ATL2Text file.

Currently, There are two versions of our mapping tool, one with AMMA translators and 
one with translators written by hand (in Java), which we got from a preliminary development 
phase of the system. Comparing both approaches, let us make the following concluding remarks.

\begin{itemize}

\item The development of hand-written translators is in general a hard task. 
Their creation, modification and reuse require to have a deep insight in the code 
and in the architecture of the platform, even more if they have a specific and/or 
complex design. For instance, the developer may be forced to directly use
lexers and parsers, or a given library which provides specific methods to generates 
the target files.

\item The development of AMMA translators does not require advanced language implementation 
skills. We show that the use of KM3 and ATL is not really a hard task. Moreover, AMMA
is supported by a set of tools~\cite{EclipseM2M} which provide a great framework to 
create and manipulate KM3, ATL and TCS models, and also for project handling.   
An independent definition of syntax concerns (ATL2Text) from metamodel concepts (KM3) is 
another advantage which gives us a more organized view that facilitates the creation and 
reuse of translators.

\item The development of hand-written translators requires more code lines. In our 
implementation, the source files of Java translators are approximately 
60\% bigger than the AMMA translators source files (ATL+KM3).

\end{itemize}

\subsection{Direct code generation}\label{sec:simpler-process}

\begin{figure}[!hbtp]
\begin{center}
\includegraphics[width=0.8\linewidth]{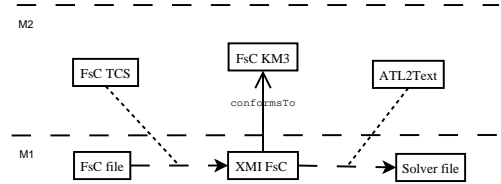}
\caption{Direct code generation.\label{fig:simpler}}
\end{center}
\end{figure}

There is another approach to develop translators using the AMMA platform. 
For instance, if we want to use just the \flatsCOMMA{} features that are 
supported by the solver, we can omit the transformation rules and we can apply 
the ATL2Text directly on the source metamodel. Figure~\ref{fig:simpler} shows 
this direct code generation process.

Although this approach is simpler, it is less flexible since we lose
the possibility of using interesting rules transformations such as the set matrix 
decomposition explained in Section~\ref{sec:ATL}.

\section{Experiments}\label{sec:bench}

We have carried out a set of tests in order to first compare the performance of 
AMMA translators (using transformation rules)
with translators written by hand, and second, to show that the automatic generation of 
solver files does not lead to a loss of performance in terms of solving time. Tests
have been performed on a 3GHz Pentium 4 with 1GB RAM running Ubuntu 6.06, and 
benchmarks used are the following~\cite{s-COMMA}:
\vspace*{-2mm}
\begin{itemize}

\item Send: The cryptoarithmetic puzzle Send + More = Money.
\vspace*{-2mm}
\item Stable: The stable marriage problem presented.
\vspace*{-2mm}
\item Queens: The N-Queens problem (n=10 and n=18).
\vspace*{-2mm}
\item Packing: Packing 8 squares into a square of area 25.
\vspace*{-2mm}
\item Production: A production-optimization problem.
\vspace*{-2mm}
\item Ineq20: 20 Linear Inequalities.
\vspace*{-2mm}
\item Engine: The assembly of a car engine subject to design constraints.
\vspace*{-2mm}
\item Sudoku: The Sudoku logic-based number placement puzzle.
\vspace*{-2mm}
\item Golfers: To schedule a golf tournament.

\end{itemize}

\begin{table}[!hbtp]
\caption{Translation times (seconds)\label{table:translators}}
\begin{center}
\begin{small}
\begin{tabular}{|l|l|ll|ll|}
\hline
 &sC to&\multicolumn{2}{l|}{FsC to Gecode/J}&\multicolumn{2}{l|}{FsC to \Eclipse{}}\\
\cline{3-6}
Benchmark&FsC&Java&AMMA&Java&AMMA\\
\hline
\hline
Send             &0.237  &0.052   &0.688   &0.048  &0.644 \\
\hline
Stable           &0.514  &0.137   &1.371   &0.143  &1.386 \\
\hline
10-Queens        &0.409  &0.106   &1.301   &0.115  &1.202 \\
\hline
18-Queens        &0.659  &1.122   &3.194   &0.272  &2.889 \\
\hline
Packing          &0.333  &0.172   &1.224   &0.133  &1.246 \\
\hline
Production       &0.288  &0.071   &0.887   &0.066  &0.783  \\
\hline
20 Ineq.         &0.343  &0.072   &0.895   &0.072  &0.891 \\
\hline
Engine           &0.285  &0.071   &0.815   &0.071  &0.844 \\
\hline
Sudoku           &3.503  &1.290   &4.924   &0.386  &4.196 \\
\hline
Golfers          &0.380  &0.098   &1.166   &0.111  &1.136 \\
\hline
\end{tabular}
\end{small}
\end{center}
\end{table}

Table~\ref{table:translators} shows preliminary results comparing AMMA translators
with translators written by hand (in Java). Column
3 and 4 give the translation times using Java and AMMA translators, from \flatsCOMMA{} (FsC) to 
Gecode/J and from \flatsCOMMA{} to \Eclipse{}, respectively. Translation times from \sCOMMA{} (sC)
to \flatsCOMMA{} are given for reference in column 2 (This process involves syntactic and 
semantic checking, and the transformations explained in Section~\ref{sec:flatscomma}). The results show that AMMA translators
are slower than Java translators, this is unsurprising since Java translators have been
designed specifically for \sCOMMA{}. They take as input a \flatsCOMMA{} definition and generate 
the solver file directly. The transformation process used by AMMA translators is not direct, 
it performs intermediate phases (XMI to XMI). Moreover, the AMMA tools are under continued development and many 
optimizations can be done especially on the parsing process of the source file (more than 60\% of 
the time is consumed by this process). Although our primary scope is not focused on performance, we 
expect to improve this using the next AMMA version. 

However, despite of this speed difference, we believe translation 
times using AMMA are acceptable and this loss of performance is a reasonable price to pay for using 
a generic approach.

In Table~\ref{table:modelsize} we compare the solver files generated by AMMA translators
\footnote{In the comparison, we do not consider solver files generated by Java translators. 
They do not have relevant differences compared to solver files generated by 
AMMA translators.} with native solver files version written by hand. 
The data is given in terms of \textit{solving time(seconds)/model size(tokens)}.
Results show that generated solver files are in general bigger than solver 
versions written by hand. This is explained by the loop unrolling and flattening 
composition processes presented in Section~\ref{sec:flatscomma}. However, this increase
in terms of code size does not cause a negative impact on 
the solving time. In general, generated solver versions 
are very competitive with hand-written versions.

Table~\ref{table:modelsize} also shows that Gecode/J files are bigger than 
\Eclipse{} files, this is because the Java syntax is more verbose than the \Eclipse{}
syntax.

\begin{table}
\caption{Solving times(seconds) and model sizes (number of tokens)\label{table:modelsize}}
\begin{center}
\begin{small}
\begin{tabular}{|l|ll|ll|}
\hline
 &\multicolumn{2}{l|}{Gecode/J}&\multicolumn{2}{l|}{\Eclipse{}}\\
\cline{2-5}
Benchmark&hand&AMMA&hand&AMMA\\
\hline
\hline
Send             &0.002/  &0.002/    &0.01/  &0.01/ \\
                 &590     &615       &231    &329   \\
\hline
Stable           &0.005/  &0.005/    &0.01/  &0.01/ \\
                 &1898    &8496      &1028   &4659   \\
\hline
10-Queens        &0.003/  &0.003/    &0.01/  &0.01/ \\
                 &460     &9159      &193    &1958   \\
\hline
18-Queens        &0.008/  &0.008/    &0.02/  &0.02/ \\
                 &460     &30219     &193    &6402   \\
\hline
Packing          &0.009/  &0.009/    &0.49/  &0.51/ \\
                 &663     &12037     &355    &3212   \\
\hline
Production       &0.026/  &0.028/    &0.014/ &0.014/ \\
                 &548     &1537      &342    &703   \\
\hline
20 Ineq          &13.886/ &14.652/   &10.34/ &10.26/ \\
                 &1576    &1964      &720    &751   \\
\hline
Engine           &0.012/  &0.012/    &0.01/  &0.01/ \\
                 &1710    &1818      &920    &1148   \\
\hline
Sudoku           &0.007/  &0.007/    &0.21/  &0.23/ \\
                 &1551/   &33192/    &797/   &11147/ \\
\hline
Golfers          &0.005/  &0.005/    &0.21/  &0.23/ \\
                 &618/    &4098/     &980/   &1147/ \\
\hline
\end{tabular}
\end{small}
\end{center}
\end{table}

\section{Related Work}\label{sec:related}

\sCOMMA{} is as related to solver-independent languages as object-oriented languages.
In the next paragraphs we compare our approach to languages belonging to these
groups.

\subsection{Solver-Independent Constraint Modeling}

Solver-independence in constraint modeling languages is a recent trend. Just
a few languages have been developed under this principle. One example is
MiniZinc, which is mainly a subset of constructs provided by Zinc, its syntax
is closely related to OPL and its solver-independent platform allows to translate 
models into Gecode and \Eclipse{} solver code. This model transformation is performed by 
a rule-based system called Cadmium~\cite{BrandPADL2008} which can be regarded as an extension of
Term-Rewriting (TR)~\cite{Baader1998} and Constraint Handling Rules (CHR)~\cite{FruehwirthJLP1998}. 
This process also involves an intermediate model called FlatZinc, which plays a similar role
than \flatsCOMMA{}, to facilitate the translation.

The implementation of our approach is quite different to Cadmium. While
Cadmium is supported by CHR and TR, our approach is based on standard model transformation
techniques, which we believe give us some advantages. For instance, ATL and 
KM3 are strongly supported by the model engineering community. A considerable amount of 
documentation and several examples are available at the Eclipse IDE site~\cite{EclipseM2M}. 
Tools such as Eclipse plug-ins are also available for developing and debugging applications. 
It is not less important to mention that ATL is considered as a standard solution for model 
transformation in Eclipse.

On the technical side, the Cadmium system is strongly tied to MiniZinc. This is a great
advantage since the rules operate directly on Zinc expression, so transformation rules are
often compact. However, this integration forces to merge the metamodel concepts of MiniZinc 
with the MiniZinc syntax. This property makes Cadmium programs more compact but less 
modular than our approach, where the syntax is defined independently from the metamodel 
(as we have presented in Section~\ref{sec:mapping-tool}).

Essence is another solver-independent language. Its syntax is addressed to users
with a background in discrete mathematics, this style makes Essence a specification 
language rather than a modeling language. The Essence execution platform allows to 
map specifications into \Eclipse{} and Minion solver~\cite{GentECAI96}. A model transformation
system called Conjure has been developed, but the integration of solver translators
is not its scope. Conjure takes as input an Essence specification and transform it to an 
intermediate OPL-like language called Essence'. Translators from Essence' to solver code 
are written by hand.

From a language standpoint, \sCOMMA{} is as expressive as MiniZinc and Essence, in fact
these approaches provide similar constructs and modeling features. However, a main feature 
of \sCOMMA{} that strongly differences it from aforementioned languages is the object-oriented 
framework provided and the possibility of modeling problems using a visual language.

\subsection{Object-Oriented Constraint Modeling and Visual Environments}

The capability of defining constraints in an object-oriented modeling 
language is the base of the object-oriented constraint modeling paradigm.
The first attempt in performing this combination was on the development of 
ThingLab~\cite{BorningTOPLAS1981}. This approach was designed for interactive 
graphical simulation. Objects were used to represent graphical elements and 
constraints defined the composition rules of these objects. 

COB~\cite{JayaramanPADL2002} is another object-oriented language, 
but its framework is not purely based on this paradigm. In fact, the 
language is a combination of objects, first order formulas and CLP (Constraint 
Logic Programming) predicates. A GUI tool is also provided for modeling problems
using CUML, a UML-like language. The focus of this language was the engineering 
design. Modelica~\cite{FritzsonECCOP1998} is another object-oriented 
approach for modeling problems from the engineering field, but it is 
mostly oriented towards simulation. 

Gianna~\cite{PaltrinieriCP1995} is a precursor visual environment for modeling CSP. But 
its modeling style is not object-oriented and the level of abstraction provided is 
lower than in UML-like languages. In this tool, CSPs are stated as constraint graphs 
where nodes represent the variables and the edges represent the constraints. 

Although these approaches do not have a system to plug-in new solvers 
and were developed for a specific application domain, we believe it is 
important to mention them.

It is important to clarify too, that object-oriented capabilities are also provided by 
languages such as CoJava~\cite{BrodskyCP2006}; and in libraries such as Gecode or ILOG SOLVER. 
The main difference here is that the host language provided is a programming language but not a high-level 
modeling language. As we have explained in Section~\ref{sec:intro}, advanced programming skills 
may be required to deal with these tools.

\section{Conclusions and Future Work}\label{sec:conclusion}

In this work we have presented \sCOMMA{}, an extensible MDD platform for
modeling CSPs. The whole system is composed by two main parts: A modeling 
tool and a mapping tool, which provide to the users the following three 
important facilities:
\begin{itemize}

\item A visual modeling language that combines the declarative aspects of constraint programming
      with the useful features of object-oriented languages.
      The user can state modular models in an intuitive way, where the compositional structure 
      of the problem can be easily maintained through the use of objects under constraints.

\item Models are stated independently from solver languages. Users are able 
      to design just one model and to target different solvers. This clearly facilitates 
      experimentation and benchmarking.

\item A model transformation system supported by the AMMA platform which follows the standards 
      of the software engineering field. The system allows users to plug-in new solvers 
      without writing translators by hand. 

\end{itemize}

     Currently, we do not use \sCOMMA{} as our source model, because its metamodel is quite large and
     defining generic mappings to different solver metamodels will be a serious challenge. However 
     we believe that this task will lead to an interesting future work, for instance to perform reverse 
     engineering (e.g. Gecode/J to \sCOMMA{} or \Eclipse{} to \sCOMMA{}). The use of AMMA for 
     model optimization will be useful too, for instance to eliminate redundant or useless 
     constraints. The definition of selective mappings is also an interesting task, for instance
     to decide, depending on the solver used, whether loops must be unrolled or the composition 
     must be flattened.  

\acks

We are grateful to the support of this research from the ``Pontificia Universidad Cat\'olica de 
Valpara{\'\i}so" under the grant ``Beca de Estudios B\'asica", and to Fr\'ed\'eric Jouault for his 
support on the implementation of the AMMA translators.

\bibliographystyle{plainnat}

\begin{thebibliography}{}

%

\bibitem[1]{ANTLR}
ANTLR Reference Manual, 2007.
\newblock http://www.antlr.org.

\bibitem[2]{Baader1998}
F. Baader and T. Nipkow.
\newblock Term rewriting and all that, Cambridge Univ. Press, 1998.

\bibitem[3]{BorningTOPLAS1981}
A. Borning.
\newblock The Programming Languages Aspects of ThingLab, a Constraint-Oriented Simulation Laboratory.
\newblock \emph{ACM Transactions on Programming Languages and Systems (ACM TOPLAS),} 3(4), pages 353--387, 1981.

\bibitem[4]{BrodskyCP2006}
A. Brodsky and H. Nash.
\newblock CoJava: Optimization Modeling by Nondeterministic Simulation.
\newblock In \emph{Proceedings of the 12th International Conference on Principles and Practice of 
                  Constraint Programming (CP 2006)} 3(4). LNCS, vol. 4204, pages 91--106, 2006.

\bibitem[5]{BrandPADL2008}
S. Brand, G. J. Duck, J. Puchinger and P. J. Stuckey.
\newblock Flexible, Rule-Based Constraint Model Linearisation.
\newblock In \emph{Proceedings of the 10th Symposium on Practical Aspects of Declarative Languages (PADL 2008)}. LNCS, vol. 4902, pages 68--83, 2008.


\bibitem[6]{DiazSAC00}
D. Diaz and P. Codognet.
\newblock The GNU Prolog System and its Implementation.
\newblock In \emph{Proceedings of the 2000 ACM Symposium on Applied Computing (SAC 2000)}, pages 728--732, 2000.

\bibitem[7]{EclipseM2M}
Eclipse Model-to-model transformation, 2008.\\
\newblock http://www.eclipse.org/m2m/.

\bibitem[8]{FrischIJCAI2005}
A. M. Frisch, C. Jefferson, B. Mart\'inez Hern\'andez and I. Miguel.
\newblock The Rules of Constraint Modelling.
\newblock In \emph{Proceedings of the 19th International Joint
               Conference on Artificial Intelligence (IJCAI 2005)}, pages 109--116, 2005.

\bibitem[9]{FrischIJCAI2007}
A. M. Frisch, M. Grum, C. Jefferson, B. Mart\'inez Hern\'andez and I. Miguel.
\newblock The Design of ESSENCE: A Constraint Language for Specifying Combinatorial Problems.
\newblock In \emph{Proceedings of the 20th International Joint
               Conference on Artificial Intelligence (IJCAI 2007)}, pages 80--87, 2007.

\bibitem[10]{FritzsonECCOP1998}
P. Fritzson and V. Engelson.
\newblock Modelica -- A Unified Object-Oriented Language for System Modeling and Simulation.
\newblock In \emph{Proceedings of the 12th European Conference on Object-Oriented Programming (ECOOP 1998)}. LNCS, vol. 1445, pages 67--90, 1998.

\bibitem[11]{FruehwirthJLP1998}
T. W. Fr{\"u}hwirth.
\newblock Theory and Practice of Constraint Handling Rules.
\newblock \emph{Journal of Logic Programming} 37(1-3), pages 95--138, 1998.

\bibitem[12]{Gecode}
Gecode System, 2006.
\newblock http://www.gecode.org.


\bibitem[13]{GentECAI96}
I. P. Gent, C. Jefferson and I. Miguel.
\newblock Minion: A Fast Scalable Constraint Solver.
\newblock In \emph{Proceedings of the 17th European Conference on Artificial Intelligence (ECAI 2006)}, pages 98--102, 2006.

\bibitem[14]{GranvilliersACM2006}
L. Granvilliers and F. Benhamou.
\newblock Algorithm 852: RealPaver: an interval solver using constraint satisfaction techniques.
\newblock \emph{ACM Transactions on Mathematical Software (ACM TOMS),} 32(1), pages 138--156, 2006.

\bibitem[15]{JaffarPOPL1987}
J. Jaffar and J.-L. Lassez.
\newblock Constraint Logic Programming.
\newblock In \emph{Proceedings of the 14th Annual ACM Symposium on Principles of Programming Languages (POPL 1987)}, pages 111--119, 1987.

\bibitem[16]{JayaramanPADL2002}
B. Jayaraman and P. Tambay.
\newblock Modeling Engineering Structures with Constrained Objects.
\newblock In \emph{Proceedings of the 4th Symposium on Practical Aspects of Declarative Languages (PADL 2002)}. LNCS, vol. 2257, pages 28--46, 2002.

\bibitem[17]{Jouault2006ATL}
F. Jouault and I. Kurtev.
\newblock Transforming Models with ATL.
\newblock In \emph{Proceedings of Satellite Events at the 8th International 
Conference on Model Driven Engineering Languages and Systems (MoDELS Satellite Events 2005)}. LNCS, vol. 3844, pages 128--138, 2005.

\bibitem[18]{Jouault2006KM3}
F. Jouault and J. B{\'e}zivin.
\newblock KM3: A DSL for Metamodel Specification.
\newblock In \emph{Proceedings of the 8th IFIP WG 6.1 International Conferenceon 
                   Formal Methods for Open Object-Based Distributed Systems (FMOODS 2006)}. LNCS, vol. 4037, pages 171--185, 2006.

\bibitem[19]{Jouault2006TCS}
F. Jouault, J. B{\'e}zivin and I. Kurtev
\newblock KM3: A DSL for Metamodel Specification.
\newblock In \emph{Proceedings of the 5th International Conference on Generative Programming and Component Engineering (GPCE 2006)}, pages 249--254, 2006.


\bibitem[20]{Kurtev2006}
I. Kurtev, J. B{\'e}zivin, F. Jouault and P. Valduriez.
\newblock Model-based DSL frameworks.
\newblock In \emph{Proceedings of Companion to the 21th Annual ACM SIGPLAN Conference on Object-Oriented Programming, Systems, Languages, and Applications (OOPSLA Companion 2006)}, pages 602--616, 2006.


\bibitem[21]{Nethercote2007}
N. Nethercote, P. J. Stuckey, R. Becket, S. Brand, G. J. Duck and G. Tack.
\newblock MiniZinc: Towards a Standard CP Modelling Language.
\newblock In \emph{Proceedings of the 13th International Conference on Principles and Practice of 
                  Constraint Programming (CP 2007)} 3(4). LNCS, vol. 4741, pages 529--543, 2007.

\bibitem[22]{OMG_OCL}
OMG - Object Constraint Language (OCL) 2.0, 2006.\\
\newblock http://www.omg.org/cgi-bin/doc?formal/2006-05-01

\bibitem[23]{OMG_UML}
OMG - The Unified Modeling Language (UML) 2.1.1 Infrastructure Specification, 2007.
\newblock http://www.omg.org/spec/UML/2.1.2/.

\bibitem[24]{OMG_MDA}
OMG - Model Driven Architecture (MDA) Guide V1.0.1, 2003
\newblock http://www.omg.org/cgi-bin/doc?omg/03-06-01.

\bibitem[25]{PaltrinieriCP1995}
M. Paltrinieri.
\newblock A Visual Constraint-Programming Environment.
\newblock In \emph{Proceedings of the 1st International Conference on Principles and Practice of 
                  Constraint Programming (CP 1995)} 3(4). LNCS, vol. 976, pages 499--514, 1995.

\bibitem[26]{PugetSCIS1994}
J.-F. Puget.
\newblock A C++ implementation of CLP.
\newblock In \emph{Proceedings of the Second Singapore International Conference on Intelligent Systems (SCIS 1994)}, 1994.


\bibitem[27]{RafehPADL07}
R. Rafeh, M. J. Garc\'{\i}a de la Banda, K. Marriott and M. Wallace.
\newblock From Zinc to Design Model.
\newblock In \emph{Proceedings of the 9th Symposium on Practical Aspects of Declarative Languages (PADL 2007)}. LNCS, vol. 4354, pages 215--229, 2007.



\bibitem[28]{s-COMMA}
s-COMMA System, 2008.
\newblock http://www.inf.ucv.cl/\~{ }rsoto/s-comma.

\bibitem[29]{SotoICTAI07}
R. Soto and L. Granvilliers.
\newblock The Design of COMMA: An Extensible Framework for Mapping Constrained Objects to Native Solver Models.
\newblock In \emph{Proceedings of the 19th IEEE International Conference on Tools with Artificial
               Intelligence (ICTAI 2007)}, pages 243--250, 2007.


\bibitem[30]{VanHentenryckBook1999}
P. Van Hentenryck.
\newblock The OPL Language.
\newblock MIT Press, 1999.

\bibitem[31]{Wallace97eclipse}
M. Wallace, S. Novello and J. Schimpf.
\newblock Technical report, IC-Parc, Imperial College, London, 1997.


\end{thebibliography}

\end{document}